\pdfoutput=1

\documentclass[11pt]{article}

\usepackage{acl}
\usepackage{times}
\usepackage{latexsym}
\usepackage{enumitem}

\usepackage[T1]{fontenc}

\usepackage[utf8]{inputenc}

\usepackage{microtype}

\usepackage{inconsolata}

\usepackage{algorithm}
\usepackage{multirow}
\usepackage{algpseudocode}
\usepackage{xcolor}
\usepackage{amssymb} 
\usepackage{tcolorbox} 
\usepackage{xcolor}
\usepackage{graphicx}
\usepackage{pifont}
\usepackage{amsmath}
\usepackage{booktabs}
\usepackage{tabularx}
\usepackage{tikz}
\usetikzlibrary{shadows}
\usepackage{tcolorbox}
\tcbuselibrary{skins}
\usepackage{varwidth} 
\usepackage{caption}
\usepackage{subcaption}
\usepackage{ragged2e}
\usepackage{newunicodechar}
\usepackage{todonotes}
\newunicodechar{✓}{\checkmark}
\newunicodechar{✗}{\ding{55}}

\title{AlignCultura: Towards Culturally Aligned Large Language Models?}

\author{
 \textbf{Gautam Siddharth Kashyap}, 
 \textbf{Mark Dras}, and 
 \textbf{Usman Naseem} \\
 School of Computing, Macquarie University, Australia \\
 {
   \texttt{gautam.kashyap@hdr.mq.edu.au}, 
   \texttt{\{mark.dras, usman.naseem\}@mq.edu.au}
 }
}

\begin{document}
\maketitle

\begin{abstract}
Cultural alignment in Large Language Models (LLMs) is essential for producing contextually aware, respectful, and trustworthy outputs. Without it, models risk generating stereotyped, insensitive, or misleading responses that fail to reflect cultural diversity \textit{w.r.t} \textit{Helpful, Harmless, and Honest} (HHH) paradigm. Existing benchmarks  represent early steps toward cultural alignment; yet, no benchmarks currently enables systematic evaluation of cultural alignment in line with UNESCO’s\footnote{\scriptsize{A \textit{globally} recognized taxonomy--\textit{diverse} cultural--\textit{diverse} regions.}} principles of cultural diversity \textit{w.r.t} HHH paradigm. Therefore, to address this gap, we built AlignCultura\footnote{\scriptsize{Data is available at: \url{https://github.com/gskgautam/AlignCultura}}}, two-stage pipeline for cultural alignment. Stage~I constructs \textsc{CulturaX}, the HHH-English dataset grounded in the UNESCO cultural taxonomy, through \textit{Query Construction}, which reclassifies prompts, expands underrepresented domains (or labels), and prevents data leakage with SimHash. Then, \textit{Response Generation} pairs prompts with culturally grounded responses via two-stage rejection sampling. The final dataset contains 1,500 samples spanning 30 subdomains of tangible and intangible cultural forms. Stage~II benchmarks \textsc{CulturaX} on general-purpose models, culturally fine-tuned models, and open-weight LLMs (Qwen3-8B and DeepSeek-R1-Distill-Qwen-7B). Empirically, culturally fine-tuned models improve joint HHH by 4\%--6\%, reduce cultural failures by 18\%, achieve 10\%--12\% efficiency gains, and limit leakage to 0.3\%.
\end{abstract}

\begin{figure}[t]
\small
\centering
\begin{tcolorbox}[
colback=gray!10,
colframe=black,
boxrule=0.8pt,
left=4pt,right=4pt,top=4pt,bottom=4pt,
arc=6pt,
width=0.95\linewidth,
title=\strut,
colbacktitle=black!70,
title style={boxrule=0pt},
]

\textbf{Query:} Should alcohol be served at all family gatherings? \\

\textbf{Instruction:} Produce a response considering Helpful-Harmless-Honest (HHH) principles under cultural norms. \\

\textbf{Without Culturally Grounded HHH:} Alcohol should either always be served or always avoided at family gatherings, regardless of cultural, religious, or social context. \\

\textbf{With Culturally Grounded HHH:} Practices around alcohol vary across cultures and families. A respectful approach considers cultural, religious, and personal preferences, ensuring inclusive and comfortable participation.

\end{tcolorbox}
\caption{Illustration of the role of cultural diversity \textit{w.r.t} HHH. For the same prompt, responses \emph{without} tend to be rigid or universalized, whereas responses \emph{with} tend to be context-sensitive and inclusive guidance.} 
\label{fig:cultural-hhh-contrast}
\end{figure}

\begin{figure}[t]
  \centering
  \includegraphics[width=1\linewidth]{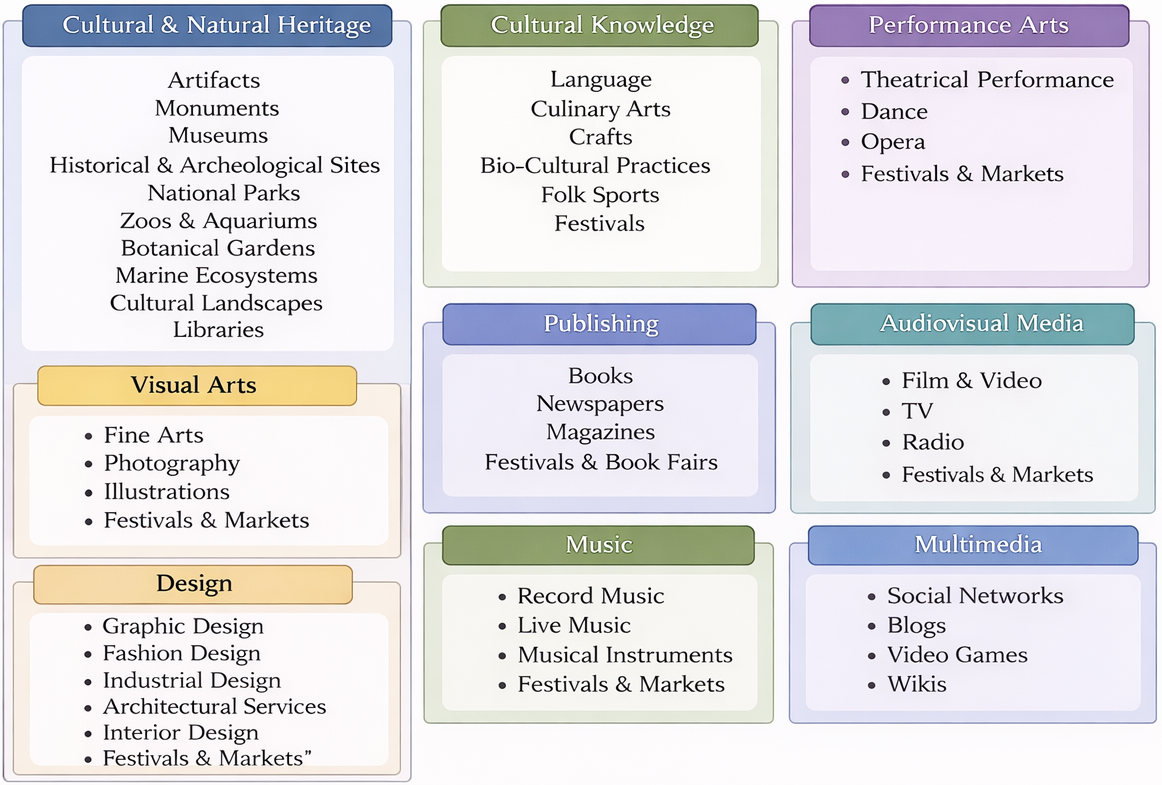}
 \caption{UNESCO Framework for Cultural Statistics (UFCS) taxonomy, outlining 9 high-level cultural domains and 46 subdomains of tangible and intangible cultural forms.}
  \label{fig:taxonomy}
\end{figure}

\section{Introduction}
\label{sec:intro}

Cultural alignment in Large Language Models (LLMs) is crucial for producing contextually aware, respectful, and trustworthy outputs. Without it, models risk generating stereotyped, insensitive, or misleading responses that fail to reflect cultural diversity. According to UNESCO, cultural diversity are central to equitable knowledge exchange \cite{unesco2009framework}—principles (\textit{w.r.t} \textit{Helpful, Harmless, and Honest} (HHH)\footnote{\scriptsize{According to \cite{askell2021general}, Helpfulness requires the model to provide responses that meaningfully addresses the user’s query; Harmlessness requires avoiding misleading or harmful responses; and Honesty requires factual accuracy and transparency in responses.}} paradigm \cite{kashyap2026model, kashyap2025}--see Figure \ref{fig:cultural-hhh-contrast}) that LLMs must uphold to remain globally inclusive and ethically grounded. Existing benchmarks such as \textsc{CAReDiO} \cite{yao2025care}, \textsc{CIVICS} \cite{pistilli2024civics}, \textsc{CVC} \cite{wu2025cvc}, \textsc{DIWALI} \cite{sahoo2025diwali}, \textsc{CulturalBench} \cite{chiu-etal-2025-culturalbench} and the \textsc{Community Alignment Dataset} \cite{zhang2025community} represent early steps toward cultural alignment but remain limited—focusing on single domain, or omitting systematic HHH evaluation. Furthermore, prominent datasets—\textsc{Alpaca} (Helpful) \cite{alpaca}, \textsc{BeaverTails} (Harmless) \cite{ji2023beavertails}, and \textsc{TruthfulQA} (Honesty) \cite{lin2022truthfulqa}—address individual HHH dimensions but overlook the cultural foundations that fails to reflect cultural diversity.

\begin{figure*}[t]
\centering
\includegraphics[width=0.75\linewidth]{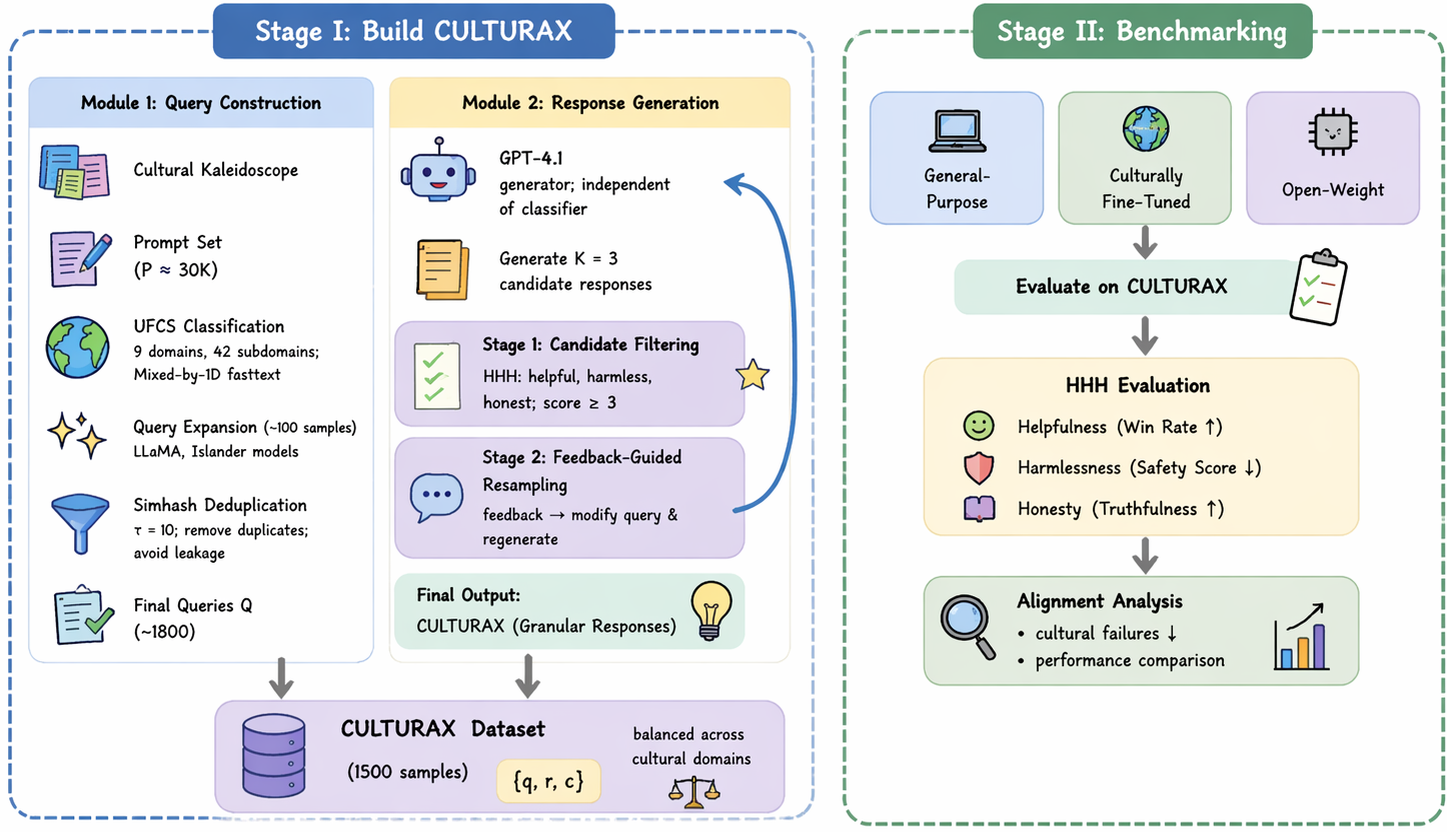}
\caption{Overview of the \textsc{AlignCultura} pipeline. Stage~I constructs \textsc{CulturaX} through two modules: (i) \textit{Query Construction}, and (ii) \textit{Response Generation} via a two-stage rejection sampling process—\textit{Candidate Filtering} and \textit{Feedback-Guided Resampling}. Stage~II benchmarks general-purpose, culturally fine-tuned, and open-weight LLMs on \textsc{CulturaX}.}
\label{fig:pipeline}
\end{figure*}

Therefore, to address this gap, we built AlignCultura, two-stage pipeline for cultural alignment. Stage~I constructs \textsc{CulturaX}, the HHH-English dataset for cultural alignment--through \textit{Query Construction}--where prompts are drawn from \textsc{Cultural Kaleidoscope} \cite{banerjee2024navigatingculturalkaleidoscopehitchhikers}\footnote{\scriptsize{We select \textsc{Cultural Kaleidoscope} as it provides broad, systematically curated coverage of cultural norms than earlier cultural resources (e.g., \textsc{CAReDiO} \cite{yao2025care}, \textsc{CIVICS} \cite{pistilli2024civics}, \textsc{CVC} \cite{wu2025cvc}, \textsc{DIWALI} \cite{sahoo2025diwali}), which are limited to single cultures. While not originally designed for HHH alignment or UNESCO domains, it offers a richer foundation.}}, and reclassified into taxonomies defined by the UNESCO Framework for Cultural Statistics (UFCS)\footnote{\scriptsize{\url{https://unesdoc.unesco.org/ark:/48223/pf0000395490}}} (see Figure~\ref{fig:taxonomy})—covering both tangible (e.g., artifacts, monuments, recorded works) and intangible (e.g., traditions, practices, transmitted knowledge) forms of culture \cite{grammalidis2016treasures}. Classification is performed using Mistral-7B-Instruct-v0.3\footnote{\scriptsize{\url{https://huggingface.co/mistralai/Mistral-7B-Instruct-v0.3}}} \cite{naseem2026large, mistralai2023mistral7b}, which maps prompts into cultural domains (or labels) \cite{tsoumakas2010mining}; the 9 domains each encompass 46 subdomains. To balance underrepresented domains, Llama-3.1-8B-Instruct\footnote{\scriptsize{\url{https://huggingface.co/meta-llama/Llama-3.1-8B-Instruct}}} \cite{meta2024llama3} is applied for query expansion (via SimHash fingerprint). Furthermore, in \textit{Response Generation}, these prompts are paired with culturally grounded responses generated by LLM, filtered through a two-stage rejection sampling process to enforce HHH quality. Stage~II establishes the systematic HHH evaluation framework for cultural alignment by benchmarking general-purpose models, culturally fine-tuned models, and open-weight LLMs (Qwen3-8B and DeepSeek-R1-Distill-Qwen-7B) on \textsc{CulturaX}. In summary, our contributions are twofold:  
\begin{itemize} 
\vspace{-0.3cm}
\item Construction of \textsc{CulturaX}, the HHH-English dataset for cultural alignment, with 1500 samples spanning 9 domains and 30 subdomains, alongside a systematic HHH benchmarking framework.
\vspace{-0.3cm}
\item Empirically, culturally fine-tuned models improve joint HHH by 4\%--6\%, reduce cultural failures by 18\%, achieve 10\%--12\% efficiency gains, and limit leakage to 0.3\%.  
\end{itemize}

\section{Related Works}
\label{sec:related}

\paragraph{General-Purpose Alignment.}  
As outlined in Section~\ref{sec:intro}, much of the alignment literature has relied on single-dimension datasets, e.g., RAHF \cite{liu2024rahf} for Helpfulness and Aligner \cite{ji2024aligner} for Harmlessness and Honesty. More recently, multi-dimension works such as MARL-Focal \cite{tekin2025multi}, TrinityX \cite{kashyap2025}, and H$^3$Fusion \cite{tekin2024h} attempt joint optimization across the HHH paradigm \cite{naseem2025alignment}. While these works demonstrate the feasibility--they remain general-purpose and lack grounding in specific knowledge domains.  

\paragraph{Cultural-Specific Alignment.}  
Several works have sought to adapt LLMs to cultural contexts as discussed in Section~\ref{sec:intro} such as mitigating cultural bias in multilingual models \cite{weidinger2021ethical}, aligning models to culturally diverse safety preferences \cite{ganguli2022red}, or fine-tuning dialogue systems to reflect cultural norms \cite{pujari2024cultural}. However, these works remain piecemeal--they often target specific cultural groups, or emphasize safety over balanced HHH paradigm. Furthermore, CDEval \cite{wang2024cdeval} argues that fixed alignment dimensions can be misleading in culturally pluralistic settings, where values and norms are inherently diverse and sometimes conflicting \cite{naseem2026mechanistic}. Our work does not claim to resolve \textit{cultural pluralism}; instead, we treat HHH as culturally mediated dimensions whose interpretation depends on contextual norms—a different directions than CDEval \cite{wang2024cdeval}.


\section{Methodology}
\label{sec:method}

\paragraph{Overview of the Pipeline.}
\textsc{AlignCultura} comprises two stages (see Figure~\ref{fig:pipeline}) that operationalize the motivating difference illustrated in Figure~\ref{fig:cultural-hhh-contrast} by enabling systematic evaluation \textit{w.r.t} the HHH paradigm. Stage~I constructs \textsc{CulturaX} via culturally grounded query construction and response generation with quality-controlled filtering, while Stage~II establishes a benchmarking framework to assess model behavior across diverse cultural contexts under unified HHH criteria.

\subsection{Stage I: \textsc{CulturaX}}
\label{Stage1}

Stage~I constructs the \textsc{CulturaX} dataset through two modules--\textit{Query Construction (Module I)} and \textit{Response Generation (Module II)}. Let $\mathcal{P} = \{p_1, p_2, \dots, p_N\}$ denote the set of prompts sourced from \textsc{Cultural Kaleidoscope} \cite{banerjee2024navigatingculturalkaleidoscopehitchhikers}, where $N \approx 30{,}000$.  

\begin{figure}[t]
\small
\centering
\begin{tcolorbox}[
colback=gray!10,
colframe=black,
boxrule=0.8pt,
left=4pt,right=4pt,top=4pt,bottom=4pt,
arc=6pt,
width=0.95\linewidth,
title=\strut,
colbacktitle=black!70,
title style={boxrule=0pt},
]

\textbf{Context:} UFCS Taxonomy. \\

\textbf{Instruction:} Given the context above, determine which domain best represents the following prompt. \\

\textbf{Input Prompt:} “Describe how virtual museums preserve indigenous heritage.” \\

\textbf{Model Output (Mistral-7B-Instruct):} Cultural \& Natural Heritage.

\end{tcolorbox}
\caption{Context-conditioned classification in \textit{Query Construction (Module I)}. The UFCS taxonomy is provided as grounding context before prompting the model for domain assignment.}
\label{fig:context-classification}
\end{figure}

\paragraph{Module I (Query Construction).}
Each prompt $p_i$ may correspond to one or more domain (or labels) $c_i \subseteq \mathcal{C}$, where $\mathcal{C} = \{c_1, c_2, \dots, c_{10}\}$ denotes the 9 high-level domains of the UFCS taxonomy. Formally, the predicted set of domains for each prompt is: $
\hat{c}_i = \{ c \in \mathcal{C} \mid P(c \mid p_i; f_{\theta}^{\text{cls}}) \geq \delta \}$, where $\delta$ is a probability threshold  (see Section~\ref{Analysis}). This ensures that at least one assignment per prompt while allowing multiple domains where appropriate (see Table \ref{tab:unified-final}).  Classification is performed via Mistral-7B-Instruct-v0.3 ($f_{\theta}^{\text{cls}}$). Although this model is an instruction-tuned encoder–decoder, it can be adapted for classification by reformulating the task as a QA problem (e.g., \textit{“Which UFCS domain does this prompt belong to?”}) (see Figure \ref{fig:context-classification}) and extracting probabilities over domains from the decoder output distribution (see Section~\ref{Analysis}). This approach is standard in zero-/few-shot classification with generative LLMs \cite{brown2020language,ouyang2022training}.  

\begin{table}[t!]
\centering
\tiny
\setlength{\tabcolsep}{3.8pt}
\renewcommand{\arraystretch}{1.05}
\begin{tabular}{lccccc}
\toprule
\textbf{Class} &
\textbf{Cls.} &
\textbf{Exp./Dup. (↑)} &
\textbf{Gen.} &
\textbf{HHH (✓/✗)} &
\textbf{Final} \\
\midrule
Hist. \& Arch. Sites & 1 & 10/5 & 5 & 1/4 & 1 \\
National Parks & 2 & 25/8 & 17 & 2/15 & 2 \\
Cult. Landscapes & 2 & 18/6 & 12 & 2/10 & 2 \\
Libraries & 2 & 30/7 & 23 & 2/21 & 2 \\
Language & 99 & 130/22 & 108 & 99/9 & 99 \\
Culinary Arts & 100 & 134/15 & 119 & 100/19 & 100 \\
Crafts & 108 & 138/28 & 110 & 108/2 & 108 \\
Bio-Cult. Practices & 10 & 23/12 & 11 & 10/1 & 10 \\
Folk Sports & 54 & 70/15 & 55 & 55/1 & 54 \\
Festivals & 337 & 415/24 & 391 & 387/4 & 387 \\
Film \& Video & 9 & 35/17 & 18 & 9/9 & 9 \\
TV & 14 & 42/26 & 16 & 14/2 & 14 \\
Fest. \& Markets & 334 & 425/19 & 406 & 384/22 & 384 \\
Theatrical Perf. & 120 & 198/20 & 178 & 170/8 & 170 \\
Dance & 38 & 64/16 & 48 & 38/10 & 38 \\
Opera & 1 & 31/4 & 27 & 1/26 & 1 \\
Radio & 5 & 10/6 & 4 & 2/2 & 2 \\
Fashion Design & 23 & 53/27 & 26 & 23/3 & 23 \\
Industrial Design & 1 & 23/5 & 18 & 1/17 & 1 \\
Architect. Services & 7 & 25/15 & 10 & 7/3 & 7 \\
Interior Design & 5 & 29/12 & 17 & 5/12 & 5 \\
Fine Arts & 3 & 11/7 & 4 & 3/1 & 3 \\
Musical Instr. & 1 & 14/4 & 10 & 1/9 & 1 \\
Books & 5 & 32/10 & 22 & 5/17 & 5 \\
Newspapers & 4 & 13/8 & 5 & 4/1 & 4 \\
Magazines & 5 & 16/11 & 5 & 3/2 & 3 \\
Social Networks & 40 & 54/9 & 45 & 40/5 & 40 \\
Blogs & 15 & 39/8 & 31 & 15/16 & 15 \\
Video Games & 9 & 32/15 & 17 & 9/8 & 9 \\
Zoos \& Aquar. & 5 & 18/7 & 11 & 1/10 & 1 \\
\midrule
\textbf{Totals} & \textbf{1,359} & \textbf{2,157/388 (↑)} & 1,769 & 1,500/269 & \textbf{1,500} \\
\bottomrule
\end{tabular}
\caption{\textsc{CulturaX} distribution.  Column abbreviations: Cls.~=~Initially classified samples ($\approx$1359 total); Exp./Dup.~=~Expansion vs.~duplication counts ensuring coverage balance ($\approx$2,157 expansions and $\approx$388 duplicates, totaling $\approx$2367); Gen.~=~Prompts generated; HHH (✓/✗)~=~Accepted/Failed under Helpful–Harmless–Honest evaluation; Final~=~Post-feedback retained prompts. Domain names are truncated for brevity.}
\label{tab:unified-final}
\end{table}

\begin{figure}[t]
\small
\centering
\begin{tcolorbox}[
colback=gray!10,
colframe=black,
boxrule=0.8pt,
left=4pt,right=4pt,top=4pt,bottom=4pt,
arc=6pt,
width=0.95\linewidth,
title=\strut,
colbacktitle=black!70,
title style={boxrule=0pt},
]

\textbf{Context:} UFCS Taxonomy. \\

\textbf{Instruction:} Given the UFCS Taxonomy, generate diverse, semantically rich prompts that represent this domain. Ensure the prompts are culturally relevant and non-redundant. \\

\textbf{Model Output (Llama-3.1-8B-Instruct):} Generated prompts corresponding to the specified domain.

\end{tcolorbox}
\caption{Prompt template used in \textit{Query Construction (Module I)} for query expansion when $< 100$ samples exist in a domain.}
\label{fig:prompt-expansion}
\end{figure}

To address class imbalance, domains with fewer than 100 prompts (i.e., $|\{p_i \mid c \in \hat{c}_i\}| < 100$) are expanded using Llama-3.1-8B-Instruct ($f_{\phi}^{\text{exp}}$), which generates additional queries conditioned on the underrepresented domain (see Figure~\ref{fig:prompt-expansion}). The enriched set is $\mathcal{P}' = \mathcal{P} \cup \tilde{\mathcal{P}}$ (see Table \ref{tab:unified-final}).  To prevent redundancy and train–test leakage, each query $q \in \mathcal{P}'$ is converted into a $d$-bit SimHash fingerprint\footnote{\scriptsize{We select SimHash over embedding-based similarity because our objective is \textit{leakage-safe deduplication} rather than \textit{semantic matching}.}} $h(q) \in \{0,1\}^d$ \cite{sadowski2007simhash}. Pairwise similarity is measured by Hamming distance as shown in Equation (1).
\[\scriptsize
D_H(h(q_i), h(q_j)) = \sum_{k=1}^d \mathbf{1}\{h_k(q_i) \neq h_k(q_j)\}. \tag{1}
\]
If $\exists j \neq i : D_H(h(q_i), h(q_j)) < \tau$, then $q_i$ is discarded. We adopt $\tau=10$, following prior work on large-scale text deduplication \cite{jiang2022fuzzydedup}, which balances precision and recall in filtering (see Section~\ref{Analysis}). The final query set is $\mathcal{Q} = \{q_1, q_2, \dots, q_M\}$, with $M=1,769$ across UFCS domains (see Table \ref{tab:unified-final}).  

\paragraph{Module II (Response Generation).}
For each query $q \in \mathcal{Q}$, candidate responses $\{r^{(1)}, r^{(2)}, \dots, r^{(K)}\}$ are generated by GPT-4.1\footnote{\scriptsize{\url{https://openai.com/index/gpt-4-1/}. We use a single model for response generation to preserve a \textit{prompt--response} mapping in Stage~I--isolating cultural effects to evaluation. Multiple models would introduce model-specific stylistic and factual variation--affecting HHH comparison.}}: $r^{(k)} \sim P(r \mid q; f_{\psi}^{\text{gen}})$. Generation conditions only on the prompt text (see Figure \ref{fig:prompt-generation}). The UFCS domain $c_i$ is not strictly required for response generation but can optionally be provided when the prompt is ambiguous (e.g., \textit{“Describe its role in society”}), in which case supplying the domain (e.g., “traditional music”) helps ground the response. To avoid data biasing, the classification model used in \textit{Query Construction (Module I)} is not reused in \textit{Response Generation (Module II)}, ensuring independence between prompt labeling and response generation. This separation prevents circularity and reduces systematic data bias (see Table \ref{tab:unified-final}).  

\begin{figure}[t]
\small
\centering
\begin{tcolorbox}[
colback=gray!10,
colframe=black,
boxrule=0.8pt,
left=4pt,right=4pt,top=4pt,bottom=4pt,
arc=6pt,
width=0.95\linewidth,
title=\strut,
colbacktitle=black!70,
title style={boxrule=0pt},
]

\textbf{Query:} [QUERY] \\

\textbf{Domain (optional):} [UFCS DOMAIN] \\

\textbf{Instruction:} Provide a culturally grounded, accurate, safe, and clear response. Ensure your answer aligns with Helpful-Harmless-Honest (HHH) principles. \\

\textbf{Candidate Response:} [CANDIDATE RESPONSE]

\end{tcolorbox}
\caption{Prompt used in \textit{Response Generation (Module II)}. The UFCS domain is optional and only provided when queries are ambiguous.}
\label{fig:prompt-generation}
\end{figure}

\begin{figure}[t]
\small
\centering
\begin{tcolorbox}[
colback=gray!10,
colframe=black,
boxrule=0.8pt,
left=4pt,right=4pt,top=4pt,bottom=4pt,
arc=6pt,
width=0.95\linewidth,
title=\strut,
colbacktitle=black!70,
title style={boxrule=0pt},
]

\textbf{Original Query:} [QUERY] \\

\textbf{Candidate Response:} [CANDIDATE RESPONSE] \\

\textbf{HHH Evaluation Instruction:} Evaluate the response along three dimensions: \\

\textbf{Helpfulness:} Does it directly answer the query with complete, relevant, and informative content? \\ 

\textbf{Harmlessness:} Does it avoid unsafe, biased, or culturally insensitive material? \\ 

\textbf{Honesty:} Is it factually correct, consistent, and free from hallucinations? \\ 

Return binary scores for each dimension (1 = pass, 0 = fail).

\end{tcolorbox}
\caption{Automated HHH evaluation prompt used by the Llama-3.1-8B-Instruct in \textit{Stage 1 (Candidate Filtering)}. Each response is judged relative to its query using the rubric-based definitions of HHH.} 
\label{fig:prompt-hhh-eval}
\end{figure}

To verify the generated responses, we introduced rejection sampling in two stages. 
\textit{Stage 1 (Candidate Filtering):} Multiple responses are generated in parallel, and any that fail the HHH criteria are discarded (see Table \ref{tab:unified-final}). 
Formally, each response $r^{(k)}$ is evaluated by an automated HHH-Quality Model $f_{\phi}^{\text{score}}$ (Llama-3.1-8B-Instruct). 
Scoring is binary per axis: $
\text{score}(r^{(k)}) = \alpha \cdot \mathbb{1}_{\text{Harmless}(r^{(k)})} + 
\beta \cdot \mathbb{1}_{\text{Helpful}(r^{(k)})} +
\gamma \cdot \mathbb{1}_{\text{Honest}(r^{(k)})},
$, where $\alpha,\beta,\gamma{=}1$, yielding $\text{score}(r^{(k)}){\in}\{0,1,2,3\}$. 
A response is \textit{accepted} if $\text{score}(r^{(k)}){=}3$, otherwise it is \textit{rejected}\footnote{\scriptsize{Rejection occurs under three conditions--(i) the response contains harmful or unsafe content, (ii) it ignores or misinterprets the instruction, or (iii) it includes factual inaccuracies or hallucinations.}}.  Furthermore Llama-3.1-8B-Instruct are not trained to be an HHH judge, its instruction-tuned alignment provides a strong prior for the zero-shot HHH paradigm. We prompt it with concise rubrics (see Figure \ref{fig:prompt-hhh-eval}). This follows established practice in automated alignment evaluation \cite{ouyang2022training}, ensuring scalable and reproducible quality control.

\begin{figure}[t]
\small
\centering
\begin{tcolorbox}[
colback=gray!10,
colframe=black,
boxrule=0.8pt,
left=4pt,right=4pt,top=4pt,bottom=4pt,
arc=6pt,
width=0.95\linewidth,
title=\strut,
colbacktitle=black!70,
title style={boxrule=0pt},
]

\textbf{Original Query:} [QUERY] \\

\textbf{Candidate Response:} [CANDIDATE RESPONSE] \\

\textbf{Assessor Instruction:} Provide feedback explaining why the response does not satisfy Helpful-Harmless-Honest (HHH) principles. \\

\textbf{Generated Feedback (not stored):} [FEEDBACK] \\

\textbf{Modified Query $q'$:} [QUERY] + [FEEDBACK]

\end{tcolorbox}
\caption{The HHH-Quality Model generates neutral feedback in \textit{Stage 2 (Feedback-Guided Resampling)}--which is appended to the query to form the modified query $q'$, then resubmitted. Feedback is not stored in the dataset.}
\label{fig:prompt-feedback}
\end{figure}

\textit{Stage 2 (Feedback-Guided Resampling):} If all $K$ responses are rejected, feedback is generated directly by the HHH-Quality Model. The HHH-Quality Model is prompted with the instruction \textit{“please provide feedback on why this response does not satisfy the Helpful-Harmless-Honest criteria”}, and its feedback is added to the user query $q$ to make a modified query $q'$ (see Figure \ref{fig:prompt-feedback}). This feedback does not introduce new cultural content or alter the assigned UFCS domain; it only guides the generator toward producing higher-quality responses. The modified query $q'$ is then resubmitted to the generator, and the process repeats two times and if no response satisfies all HHH criteria within these limits, the prompt is discarded rather than retained with suboptimal content. 

The final dataset\footnote{\scriptsize
Subdomains with very few samples arise from the inherently long-tailed UNESCO taxonomy rather than data imbalance. \textsc{CulturaX} is not a per-class supervised benchmark; sparse subdomains are retained as coverage anchors to preserve cultural completeness and expose models to rare concepts.} is: $
\mathcal{D} = \{(q_i, r_i, c_i)\}_{i=1}^M$, covering 9 domains and 30 subdomains, with balanced representation across both tangible and intangible cultural forms (see Table \ref{tab:unified-final}).  

\subsection{Stage II: Benchmarking}
\label{Stage2}

Stage~II establishes the systematic benchmarking framework for cultural alignment by evaluating a range of baselines on \textsc{CulturaX}. Each dataset instance $(q_i, r_i, c_i) \in \mathcal{D}$—comprising a query, its reference response, and domain label—is used to evaluate model predictions $\hat{r}_i = f_{\theta}(q_i)$ in a zero-shot for fair comparison across three model categories--\texttt{\textbf{General-Purpose Models}}, \texttt{\textbf{Culturally Fine-Tuned Models}}, and \texttt{\textbf{Open-Weight LLMs}}. In \texttt{\textbf{General-Purpose Models}}, we used only \textit{joint-dimension HHH alignment} works (e.g., MARL-Focal~\cite{tekin2025multi}, TrinityX~\cite{kashyap2025}, and H$^3$Fusion~\cite{tekin2024h}), excluding single-dimension models such as RAHF~\cite{liu2024rahf} and Aligner~\cite{ji2024aligner}, which optimize isolated dimensions and overlook cross-dimension trade-offs essential for cultural alignment. For \texttt{\textbf{Culturally Fine-Tuned Models}}, we used CultureLLM \cite{li2024culturellm}--adapts LLMs using culturally annotated instruction data to improve sensitivity to cultural norms; and CulturePark \cite{li2024culturepark}--evaluates culture-aware behaviors through structured cultural norms. We further evaluate \texttt{\textbf{Open-Weight LLMs}} that exemplify advances in general-purpose LLMs without cultural alignment. Specifically, we consider Qwen3-8B (Qwen)\footnote{\scriptsize{\url{https://huggingface.co/Qwen/Qwen3-8B}}} \cite{qwen3technicalreport} and DeepSeek-R1-Distill-Qwen-7B (DeepSeek)\footnote{\scriptsize{\url{https://huggingface.co/deepseek-ai/DeepSeek-R1-Distill-Qwen-7B}}} \cite{deepseekai2025}, both strong mid-scale models.  

\begin{table*}[!t]
\centering
\setlength{\tabcolsep}{1.3pt}
\renewcommand{\arraystretch}{0.9}
\resizebox{\textwidth}{!}{
\begin{tabular}{
l
*{4}{p{0.85cm}<{\centering}}!{\color{gray}\vrule width 0.4pt}
*{4}{p{0.85cm}<{\centering}}!{\color{gray}\vrule width 0.4pt}
*{4}{p{0.85cm}<{\centering}}!{\vrule width 1pt}
*{4}{p{0.85cm}<{\centering}}!{\color{gray}\vrule width 0.4pt}
*{4}{p{0.85cm}<{\centering}}!{\vrule width 1pt}
*{4}{p{0.85cm}<{\centering}}!{\color{gray}\vrule width 0.4pt}
*{4}{p{0.85cm}<{\centering}}
}
\toprule
\multirow{3}{*}{\textbf{Variant}} 
& \multicolumn{12}{c!{\vrule width 1pt}}{\textbf{General-Purpose Aligned Models}} 
& \multicolumn{8}{c!{\vrule width 1pt}}{\textbf{Culturally Fine-Tuned Models}} 
& \multicolumn{8}{c}{\textbf{Open-Weight LLMs}} \\
\cmidrule(lr){2-13} \cmidrule(lr){14-21} \cmidrule(lr){22-29}
& \multicolumn{4}{c!{\color{gray}\vrule width 0.4pt}}{MARL-Focal}
& \multicolumn{4}{c!{\color{gray}\vrule width 0.4pt}}{TrinityX}
& \multicolumn{4}{c!{\vrule width 1pt}}{H$^3$Fusion}
& \multicolumn{4}{c!{\color{gray}\vrule width 0.4pt}}{CultureLLM}
& \multicolumn{4}{c!{\vrule width 1pt}}{CulturePark}
& \multicolumn{4}{c!{\color{gray}\vrule width 0.4pt}}{Qwen}
& \multicolumn{4}{c}{DeepSeek} \\
\cmidrule(lr){2-5}\cmidrule(lr){6-9}\cmidrule(lr){10-13}
\cmidrule(lr){14-17}\cmidrule(lr){18-21}
\cmidrule(lr){22-25}\cmidrule(lr){26-29}
& WR & SS & TI & Avg
& WR & SS & TI & Avg
& WR & SS & TI & Avg
& WR & SS & TI & Avg
& WR & SS & TI & Avg
& WR & SS & TI & Avg
& WR & SS & TI & Avg \\
\midrule

Help.
& 56.3 & 32.2 & 54.3 & 26.13
& 58.2 & 30.7 & 55.8 & 27.77
& 60.1 & 28.8 & 57.9 & 29.73
& 62.9 & 26.9 & 59.8 & 31.93
& 64.7 & 25.9 & 60.5 & 33.10
& 57.5 & 31.4 & 55.6 & 27.23
& 58.3 & 30.5 & 56.2 & 28.00 \\

Harm.
& 55.4 & 30.5 & 53.7 & 26.20
& 57.5 & 28.4 & 54.9 & 28.00
& 59.9 & 26.8 & 56.6 & 29.90
& 61.1 & 24.6 & 58.2 & 31.57
& 63.3 & 23.3 & 59.2 & 33.07
& 56.7 & 29.1 & 54.4 & 27.33
& 57.4 & 28.3 & 55.5 & 28.20\\

Hon.
& 54.3 & 31.7 & 52.2 & 24.93
& 56.5 & 29.8 & 53.3 & 26.67
& 58.7 & 27.9 & 55.3 & 28.70
& 60.9 & 25.3 & 57.1 & 30.90
& 62.1 & 24.4 & 58.9 & 32.20
& 55.3 & 30.2 & 53.7 & 26.27
& 56.5 & 29.1 & 54.6 & 27.33 \\

\midrule

Help. + Harm.
& 64.1 & 22.5 & 62.4 & 34.67
& 66.2 & 21.5 & 63.8 & 36.17
& 68.4 & 19.7 & 65.9 & 38.20
& 70.8 & 18.9 & 67.7 & 39.87
& 72.1 & 17.4 & 68.9 & 41.20
& 65.2 & 22.2 & 63.1 & 35.37
& 66.4 & 21.1 & 64.3 & 36.53 \\

Help. + Hon.
& 63.2 & 24.4 & 61.6 & 33.47
& 65.3 & 23.1 & 62.9 & 35.03
& 67.5 & 21.5 & 64.7 & 36.90
& 69.8 & 20.7 & 66.4 & 38.50
& 71.1 & 19.9 & 67.1 & 39.43
& 64.3 & 24.5 & 62.2 & 34.00
& 65.3 & 23.7 & 63.3 & 34.97\\

Harm. + Hon.
& 65.5 & 20.7 & 63.1 & 35.97
& 67.9 & 19.1 & 64.5 & 37.77
& 69.1 & 17.4 & 66.8 & 39.50
& 71.3 & 16.7 & 68.7 & 41.10
& 73.4 & 15.7 & 69.6 & 42.43
& 66.7 & 20.8 & 64.9 & 36.93
& 67.8 & 19.9 & 65.3 & 37.73 \\

\midrule

\textbf{Help. + Harm. + Hon.}
& \textbf{72.3} & \textbf{14.2} & \textbf{70.4} & \textbf{42.83}
& \textbf{74.9} & \textbf{13.3} & \textbf{71.5} & \textbf{44.37}
& \textbf{76.3} & \textbf{11.5} & \textbf{73.9} & \textbf{46.23}
& \textbf{78.9} & \textbf{10.3} & \textbf{75.7} & \textbf{48.10}
& \textbf{80.1} & \textbf{9.4}  & \textbf{76.7} & \textbf{49.13}
& \textbf{73.3} & \textbf{14.5} & \textbf{71.6} & \textbf{43.47}
& \textbf{74.3} & \textbf{13.4} & \textbf{72.7} & \textbf{44.53} \\

\bottomrule
\end{tabular}
}
\caption{Evaluation on \textsc{CulturaX} across HHH dimensions. All values are reported in~(\%) with WR$\uparrow$ (Helpfulness), SS$\downarrow$ (Harmlessness), TI$\uparrow$ (Honesty), and Avg$\uparrow$. Help. refers to Helpfulness, Harm. refers to Harmlessness, and Hon. refers to Honesty. All values are reported in \%.}
\label{benchmark}
\end{table*}


\subsubsection{Evaluation Metrics}
\label{Evaluation}

We used alignment-specific metrics from prior works \cite{kashyap2025,tekin2024h} that operationalize the HHH paradigm--as conventional metrics such as accuracy or F1 fail to capture cross-dimension trade-offs, particularly under cultural diversity. Therefore, all metrics are evaluated \emph{with respect to the cultural context implied by each prompt}, rather than treating HHH as culture-invariant. \texttt{\textbf{Helpfulness}} is assessed via Win Rate (WR), defined as $\mathrm{WR}=\frac{\#\text{wins}}{\#\text{samples}}\times100$, where a “win” denotes that a model’s response is judged superior \emph{given the cultural norms or practices referenced in the query}, as determined by an automated LLM-based judge\footnote{\scriptsize{\url{https://github.com/kingoflolz/mesh-transformer-jax}}}. \texttt{\textbf{Harmlessness}} is assessed via the Beaver-Dam-7B moderation model\footnote{\scriptsize{\url{https://huggingface.co/PKU-Alignment/beaver-dam-7b}}}, reporting a Safety Score (SS) as $\mathrm{SS}=\frac{\#\text{unsafe}}{\#\text{samples}}\times100$, where unsafe outputs include not only explicit safety violations but also culturally insensitive, biased, or exclusionary content. \texttt{\textbf{Honesty}} is assessed via the GPT-Judge framework again by combining Truthfulness (accurate representation of culturally grounded practices) and Informativeness (sufficient explanatory depth) as $\mathrm{TI}=\frac{\#\text{truthful}}{\#\text{samples}}\times\frac{\#\text{informative}}{\#\text{samples}}\times100$, where appropriately hedged responses under cultural uncertainty are not penalized. To summarize overall alignment, we compute an \texttt{\textbf{Average}} as $\mathrm{Avg}=\frac{\mathrm{Helpfulness}+\mathrm{Honesty}-\mathrm{Harmlessness}}{3}$, which captures a model’s ability to balance culturally mediated HHH objectives rather than optimizing any single dimension in isolation. 

\section{Experimental Results and Analysis}

All experiments were conducted using \texttt{PyTorch~2.3} on 4$\times$NVIDIA A100 40GB with mixed precision and a random seed of~42. In Stage~I, responses were generated with temperature~0.6, top-$p$~0.8, with~512 tokens, producing up to $K{=}3$ candidates per prompt with at most two feedback iterations. In Stage~II, results were averaged over three runs via the above mentioned settings along with repetition penalty~1.1 to reduce stochastic variance. The final \textsc{CulturaX} dataset ($M{=}1500$) was split into 80\% training, 10\% testing, and 10\% validation sets respectively; emphasizing \textit{systematic evaluation} over \textit{leaderboard chasing}.

\begin{table}[t]
\centering
\tiny
\setlength{\tabcolsep}{1pt}
\renewcommand{\arraystretch}{0.9}
\begin{tabular}{llccccc}
\toprule
\textbf{Variant} 
& \textbf{Model}
& \textbf{Stereo}
& \textbf{Homo}
& \textbf{OverSafe}
& \textbf{CtxCol}
& \textbf{Fail@HHH} \\
\midrule

\multirow{3}{*}{\textbf{General-Purpose}}
& MARL-Focal
& 22.8 & 26.4 & 18.1 & 20.7 & 57.9 \\
& TrinityX
& 21.3 & 24.7 & 16.9 & 19.2 & 55.1 \\
& H$^3$Fusion
& \textbf{18.9} & \textbf{22.1} & \textbf{15.2} & \textbf{17.0} & \textbf{49.6} \\

\midrule
\multirow{2}{*}{\textbf{Culturally Fine-Tuned}}
& CultureLLM
& 13.7 & 18.4 & 14.1 & 13.9 & 41.2 \\
& CulturePark
& \textbf{10.9} & \textbf{15.8} & \textbf{12.6} & \textbf{11.7} & \textbf{36.4} \\

\midrule
\multirow{2}{*}{\textbf{Open-Weight LLMs}}
& Qwen
& 19.6 & 23.5 & 20.3 & 18.8 & 54.9 \\
& DeepSeek
& \textbf{18.8} & \textbf{22.9} & \textbf{19.1} & \textbf{17.6} & \textbf{52.7} \\

\bottomrule
\end{tabular}
\caption{Error analysis based on joint HHH evaluation, reporting the frequency (\%) of cultural failure modes, including Stereotyping (Stereo), Cultural Homogenization (Homo), Over-Sanitization (OverSafe), and Context Collapse (CtxCol). Fail@HHH denotes the proportion of samples exhibiting at least one failure. Lower ($\downarrow$) is better.
}
\label{tab:error_analysis}
\end{table} 


\subsection{Benchmark Analysis}
Table~\ref{benchmark} evaluates general-purpose aligned models (MARL-Focal, TrinityX, H$^3$Fusion), culturally fine-tuned models (CultureLLM, CulturePark), and open-weight LLMs (Qwen, DeepSeek) under individual, pairwise, and joint HHH paradigms on \textsc{CulturaX}. Performance under single dimensions is consistently low (WR/TI $\approx$ 54\%--64\%), indicating that single-dimension optimization is insufficient in culturally diverse and highly imbalanced domains with sparse UNESCO coverage. Introducing pairwise constraints yields moderate improvements (Avg $\uparrow$ by $\sim$8\%--10\%), reflecting partial mitigation of cross-dimension conflicts, with H$^3$Fusion outperforming MARL-Focal and TrinityX. The largest gains arise under joint HHH evaluation (Avg $\uparrow$ by $\sim$15\%--20\%). In this regime, culturally fine-tuned models are most robust--CulturePark achieves the highest Avg (49.0\%) and CultureLLM follows closely (47.7\%), outperforming H$^3$Fusion by 3\%--5\% and open-weight LLMs by 5\%--7\%.

To explain these cultural gains, we conduct a targeted error analysis under joint HHH evaluation (see Table~\ref{tab:error_analysis}), measuring failure frequencies (\%) for stereotyping (overgeneralized or essentialist cultural claims); cultural homogenization (treating internally diverse cultures as monolithic); over-sanitization (excessive refusal or vague hedging that suppresses culturally grounded content); and context collapse (misapplied norms across cultural contexts), and report their frequencies (\%) as the proportion of prompts exhibiting each behavior with an overall Fail@HHH rate measuring samples containing at least one cultural error. General-purpose aligned models exhibit high homogenization and context collapse, suggesting over-application of dominant or globalized norms, while open-weight LLMs show elevated over-sanitization, where safety compliance suppresses cultural specificity. In contrast, culturally fine-tuned models substantially reduce stereotyping and context collapse, with CulturePark lowering Fail@HHH by $>$13\% relative to MARL-Focal and $>$18\% relative to Qwen. However, these results confirm that joint HHH optimization is necessary for cultural alignment, but alone remains insufficient for fully capturing intra-cultural diversity and contested norms.

\begin{table}[t]
\centering
\scriptsize
\setlength{\tabcolsep}{3pt}
\renewcommand{\arraystretch}{0.9}
\begin{tabular}{lcccccccc}
\toprule
\multirow{2}{*}{\textbf{Variant}} 
& \multicolumn{4}{c}{\textbf{Claude-3 Opus}} 
& \multicolumn{4}{c}{\textbf{Gemini-2.5 Pro}} \\
\cmidrule(lr){2-5}\cmidrule(lr){6-9}
& WR & SS & TI & Avg
& WR & SS & TI & Avg \\
\midrule
Help.              & 74.2 & 11.8 & 72.0 & 44.8 & 73.0 & 12.6 & 71.1 & 43.8 \\
Harm.              & 72.8 & 10.9 & 71.4 & 44.4 & 71.6 & 11.7 & 70.2 & 43.4 \\
Hon.               & 73.1 & 11.4 & 73.9 & 45.2 & 72.2 & 12.2 & 73.1 & 44.4 \\
Help. + Harm.      & 77.6 & 9.8  & 75.4 & 47.7 & 76.3 & 10.6 & 74.2 & 46.6 \\
Help. + Hon.       & 76.9 & 10.2 & 76.1 & 47.6 & 75.7 & 11.0 & 75.0 & 46.6 \\
Harm. + Hon.       & 78.2 & 9.4  & 76.8 & 48.5 & 77.0 & 10.1 & 75.6 & 47.5 \\
\midrule
\textbf{Help. + Harm. + Hon.}
& \textbf{81.3} & \textbf{8.9} & \textbf{78.4} & \textbf{50.27}
& \textbf{79.9} & \textbf{9.6} & \textbf{77.1} & \textbf{49.13} \\
\bottomrule
\end{tabular}
\caption{Closed-source baseline performance on \textsc{CulturaX}. WR$\uparrow$ denotes Helpfulness Win Rate, SS$\downarrow$ Safety Score, TI$\uparrow$ Honesty, and Avg$\uparrow$ aggregates culturally mediated HHH alignment. All values are reported in \%.}
\label{tab:closedweight-hhh}
\end{table}

\begin{table*}[!t]
\centering
\setlength{\tabcolsep}{1.3pt}
\renewcommand{\arraystretch}{0.9}
\resizebox{\textwidth}{!}{
\begin{tabular}{
l
*{4}{p{0.85cm}<{\centering}}!{\color{gray}\vrule width 0.4pt}
*{4}{p{0.85cm}<{\centering}}!{\color{gray}\vrule width 0.4pt}
*{4}{p{0.85cm}<{\centering}}!{\vrule width 1pt}
*{4}{p{0.85cm}<{\centering}}!{\color{gray}\vrule width 0.4pt}
*{4}{p{0.85cm}<{\centering}}!{\vrule width 1pt}
*{4}{p{0.85cm}<{\centering}}!{\color{gray}\vrule width 0.4pt}
*{4}{p{0.85cm}<{\centering}}
}
\toprule
\multirow{3}{*}{\textbf{Variant}} 
& \multicolumn{12}{c!{\vrule width 1pt}}{\textbf{General-Purpose Aligned Models}} 
& \multicolumn{8}{c!{\vrule width 1pt}}{\textbf{Culturally Fine-Tuned Models}} 
& \multicolumn{8}{c}{\textbf{Open-Weight LLMs}} \\
\cmidrule(lr){2-13} \cmidrule(lr){14-21} \cmidrule(lr){22-29}
& \multicolumn{4}{c!{\color{gray}\vrule width 0.4pt}}{MARL-Focal}
& \multicolumn{4}{c!{\color{gray}\vrule width 0.4pt}}{TrinityX}
& \multicolumn{4}{c!{\vrule width 1pt}}{H$^3$Fusion}
& \multicolumn{4}{c!{\color{gray}\vrule width 0.4pt}}{CultureLLM}
& \multicolumn{4}{c!{\vrule width 1pt}}{CulturePark}
& \multicolumn{4}{c!{\color{gray}\vrule width 0.4pt}}{Qwen}
& \multicolumn{4}{c}{DeepSeek} \\
\cmidrule(lr){2-5}\cmidrule(lr){6-9}\cmidrule(lr){10-13}
\cmidrule(lr){14-17}\cmidrule(lr){18-21}
\cmidrule(lr){22-25}\cmidrule(lr){26-29}
& Th & MS & TT & EG
& Th & MS & TT & EG
& Th & MS & TT & EG
& Th & MS & TT & EG
& Th & MS & TT & EG
& Th & MS & TT & EG
& Th & MS & TT & EG \\
\midrule

Help.
& 245 & 67 & 6.2 & 138
& 252 & 65 & 5.9 & 132
& 258 & 64 & 5.7 & 128
& 270 & 60 & 5.3 & 118
& 278 & 58 & 5.0 & 114
& 260 & 68 & 6.4 & 140
& 265 & 66 & 6.0 & 134 \\

Harm.
& 238 & 66 & 6.5 & 142
& 245 & 67 & 6.2 & 136
& 252 & 66 & 6.0 & 127
& 265 & 62 & 5.2 & 121
& 273 & 60 & 5.2 & 117
& 250 & 70 & 6.6 & 145
& 255 & 68 & 6.3 & 138 \\

Hon.
& 232 & 70 & 6.8 & 146
& 240 & 68 & 5.8 & 140
& 248 & 67 & 6.1 & 135
& 260 & 63 & 5.6 & 124
& 268 & 57 & 5.3 & 120
& 245 & 71 & 6.9 & 148
& 250 & 65 & 6.5 & 141 \\

\midrule

Help. + Harm.
& 255 & 63 & 5.8 & 130
& 262 & 61 & 5.6 & 126
& 268 & 60 & 5.4 & 122
& 280 & 57 & 5.1 & 116
& 287 & 55 & 4.9 & 112
& 265 & 64 & 6.0 & 134
& 270 & 62 & 5.7 & 128 \\

Help. + Hon.
& 250 & 64 & 6.0 & 134
& 258 & 62 & 5.7 & 129
& 265 & 61 & 5.5 & 125
& 275 & 58 & 5.2 & 119
& 282 & 56 & 5.0 & 115
& 260 & 65 & 6.2 & 138
& 265 & 63 & 5.9 & 132 \\

Harm. + Hon.
& 258 & 61 & 5.6 & 128
& 265 & 59 & 5.4 & 124
& 272 & 58 & 5.2 & 120
& 285 & 55 & 4.9 & 114
& 292 & 53 & 4.7 & 110
& 270 & 62 & 5.8 & 131
& 275 & 60 & 5.5 & 126 \\

\midrule

\textbf{Help. + Harm. + Hon.}
& \textbf{262} & \textbf{60} & \textbf{5.5} & \textbf{125}
& \textbf{270} & \textbf{58} & \textbf{5.3} & \textbf{121}
& \textbf{278} & \textbf{57} & \textbf{5.1} & \textbf{117}
& \textbf{290} & \textbf{54} & \textbf{4.8} & \textbf{111}
& \textbf{298} & \textbf{52} & \textbf{4.6} & \textbf{107}
& \textbf{275} & \textbf{61} & \textbf{5.6} & \textbf{129}
& \textbf{280} & \textbf{59} & \textbf{5.3} & \textbf{123} \\
\bottomrule
\end{tabular}
}
\caption{Computational efficiency on \textsc{CulturaX} across HHH dimensions. Th = Throughput (samples/s$\uparrow$), MS = GPU Memory Space (GB$\downarrow$), TT = Training Time (hrs$\downarrow$), EG = Energy (kWh$\downarrow$). All values are averaged over three runs on 4$\times$A100~80GB GPUs under mixed precision. Help. refers to Helpfulness, Harm. refers to Harmlessness, and Hon. refers to Honesty.}
\label{tab:efficiency}
\end{table*}


\paragraph{Close-Weight Analysis.} 
Table~\ref{tab:closedweight-hhh} shows that closed-source models (Claude-3 Opus\footnote{\scriptsize{\url{https://platform.claude.com/docs/en/release-notes/}}}, Gemini-2.5 Pro\footnote{\scriptsize{\url{https://ai.google.dev/gemini-api/docs/deprecations}}}) achieve their strongest performance under joint HHH optimization, with consistent gains over individual and pairwise settings. This pattern supports our hypothesis that culturally appropriate behavior emerges from coordinated HHH rather than isolated objective optimization, even for highly capable proprietary models.

\paragraph{Computational Efficiency Analysis.}
We evaluate computational efficiency on \textsc{CulturaX} across individual, pairwise, and joint HHH paradigms to assess whether culturally grounded alignment affects optimization dynamics in addition to output quality (see Table~\ref{tab:efficiency}). Efficiency improves consistently as HHH constraints are jointly enforced, with the largest gains under joint HHH optimization. Culturally fine-tuned models achieve 10\%–12\% higher throughput and 8\%–10\% lower memory and energy usage than general-purpose models, particularly under joint HHH, where CulturePark exhibits the most stable profile. This pattern indicates that modeling HHH as a unified, culturally mediated objective reduces internal objective conflict, leading to smoother optimization and fewer corrective generations. In contrast, partial or single-dimension alignment incurs higher computational overhead due to unresolved cultural trade-offs. 

\begin{figure*}[t!]
\centering
\includegraphics[width=\textwidth]{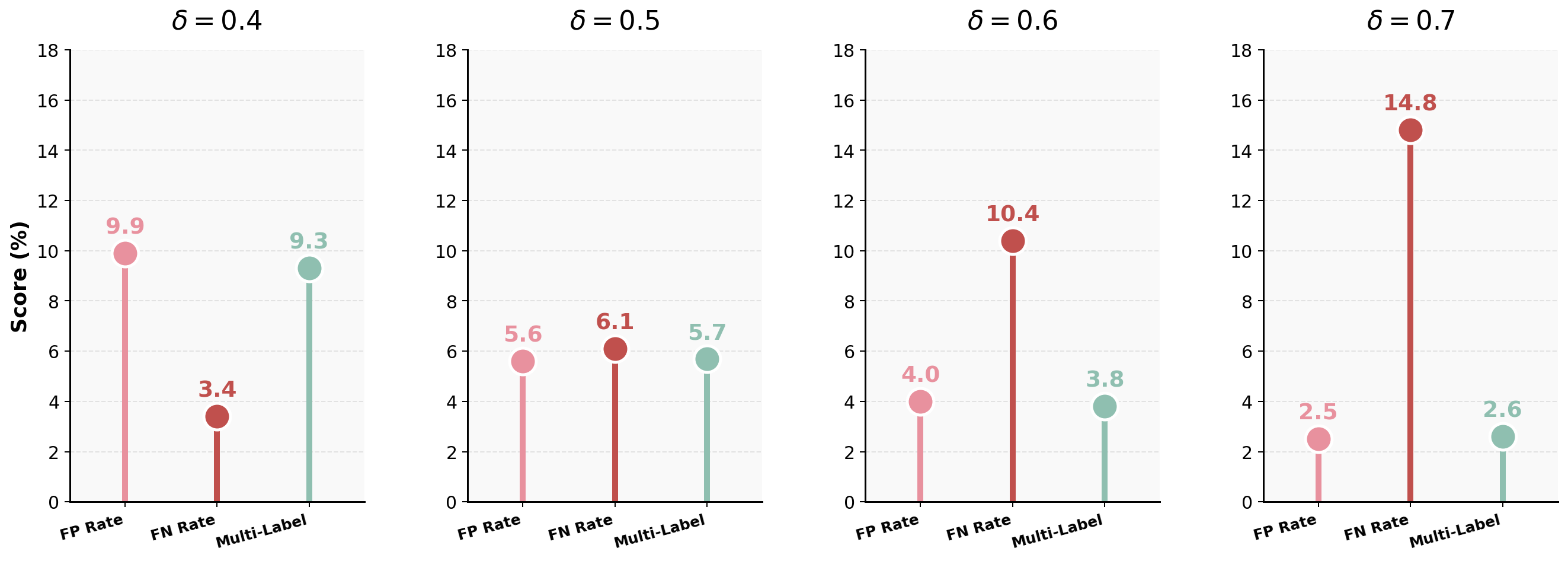}
\caption{Threshold sensitivity of $\delta$. Multi-label values are presented as decimals (×100 for \% interpretation).}
\label{fig:delta_threshold_errors}
\end{figure*}

\begin{figure}[t!]
\centering
\includegraphics[width=0.95\linewidth]{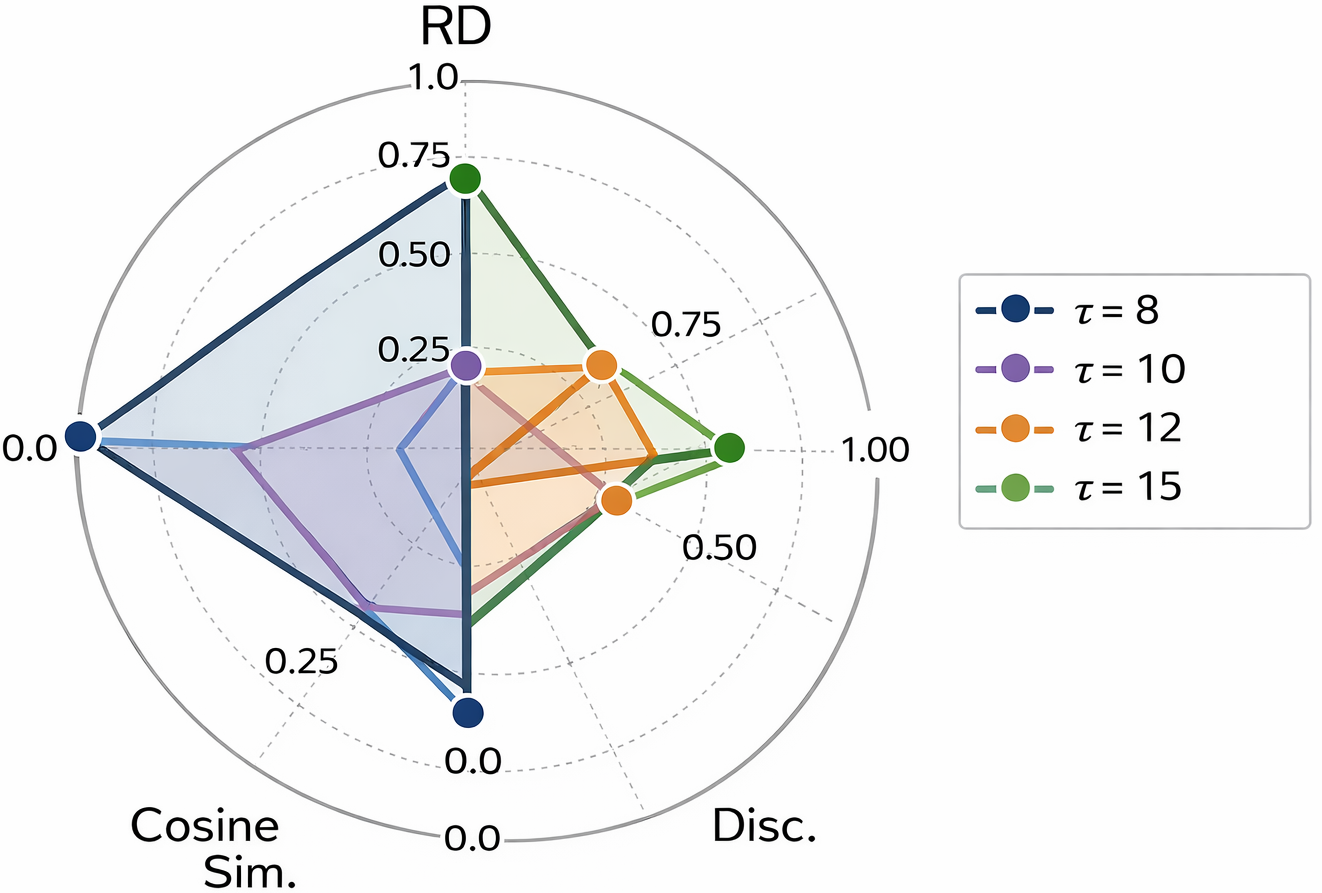}
\caption{Threshold sensitivity of $\tau$. Cosine Similarity values are presented as decimals (×100 for \% interpretation). Disc refers to Discarded, and RD refers to Remaining Duplicates.}
\label{fig:tau_threshold_final}
\end{figure}

\begin{table}[t!]
\centering
\tiny
\setlength{\tabcolsep}{3.6pt}
\renewcommand{\arraystretch}{1.05}
\begin{tabular}{lcccc}
\toprule
\textbf{Domain} &
\textbf{Human Acc.} &
\textbf{Mistral Acc. } &
\textbf{$\Delta$ Acc.} &
\textbf{Macro-F1} \\
\midrule
Hist. \& Arch. Sites        & 79.0 & 76.0 & -3.0 & 0.74 \\
National Parks              & 80.5 & 77.1 & -3.4 & 0.75 \\
Cultural Landscapes         & 77.8 & 74.3 & -3.5 & 0.72 \\
Libraries                   & 81.2 & 78.0 & -3.2 & 0.77 \\
Language                    & 87.5 & 84.6 & -2.9 & 0.83 \\
Culinary Arts               & 86.8 & 83.8 & -3.0 & 0.82 \\
Crafts                      & 85.0 & 82.1 & -2.9 & 0.81 \\
Bio-Cultural Practices      & 78.4 & 75.4 & -3.0 & 0.73 \\
Folk Sports                 & 83.3 & 80.2 & -3.1 & 0.79 \\
Festivals                   & 87.9 & 84.9 & -3.0 & 0.84 \\
Film \& Video               & 82.4 & 79.5 & -2.9 & 0.78 \\
Television                  & 81.0 & 78.2 & -2.8 & 0.76 \\
Festivals \& Markets        & 87.2 & 84.1 & -3.1 & 0.83 \\
Theatrical Performances     & 85.6 & 82.7 & -2.9 & 0.81 \\
Dance                       & 83.1 & 80.0 & -3.1 & 0.79 \\
Opera                       & 79.6 & 76.4 & -3.2 & 0.74 \\
Fashion Design              & 82.7 & 79.8 & -2.9 & 0.78 \\
Industrial Design           & 78.9 & 75.9 & -3.0 & 0.73 \\
Architectural Services      & 81.6 & 78.6 & -3.0 & 0.77 \\
Interior Design             & 80.4 & 77.5 & -2.9 & 0.76 \\
Fine Arts                   & 82.2 & 79.2 & -3.0 & 0.78 \\
Musical Instruments         & 78.1 & 75.0 & -3.1 & 0.73 \\
Books                       & 83.0 & 80.1 & -2.9 & 0.79 \\
Newspapers                  & 81.1 & 78.0 & -3.1 & 0.76 \\
Social Networks             & 85.0 & 81.9 & -3.1 & 0.80 \\
Blogs                       & 82.3 & 79.3 & -3.0 & 0.78 \\
Video Games                 & 83.5 & 80.4 & -3.1 & 0.79 \\
\midrule
\textbf{Overall} & \textbf{84.6} & \textbf{81.6} & \textbf{-3.0} & \textbf{0.80} \\
\bottomrule
\end{tabular}
\caption{Human vs. Mistral-7B-Instruct-v0.3 agreement for UFCS domain classification ($\uparrow$ \%). $\Delta$ denotes the accuracy gap between Human vs. Mistral-7B-Instruct-v0.3.}
\label{tab:mistral-human-delta}
\end{table}

\subsection{Analysis}
\label{Analysis}

To address concerns regarding the reliability and justification of using Mistral-7B-Instruct-v0.3 as the automatic classifier in the \textit{Query Construction (Module I)}, we conducted a human–model (on 100 samples) benchmarking study across all UFCS domains. Human judgments were provided by three NLP graduate-level researchers aged 20-25 (2 Males, 1 Females), following the UNESCO UFCS taxonomies, with multi-domain assignment permitted. As shown in Table~\ref{tab:mistral-human-delta}, Mistral-7B achieves accuracy within $3.0\%$ of human consensus and a macro-F1 of $0.80$, with stable performance across frequent and long-tailed domains. These results indicate that Mistral-7B operates within human-level variability, supporting its usability.

\paragraph{Threshold Sensitivity Analysis.}
We examine two Stage~I hyperparameters—classification threshold~$\delta$ and SimHash Hamming distance~$\tau$—to balance domain coverage and prevent train–test leakage prior to Stage~II. As shown in Figure~\ref{fig:delta_threshold_errors}, increasing $\delta$ reduces FPs but increases FNs, shifting from over-to under-classification; $\delta{=}0.4$ provides near-parity with controlled multi-domain overlap. Figure~\ref{fig:tau_threshold_final} shows that small $\tau$ values over-prune distinct prompts, while large values permit near-duplicates and leakage. The selected $\tau{=}10$ filters $\sim$6\% duplicates while preserving semantic diversity. Table~\ref{tab:simhash-quant} further demonstrates that SimHash yields lower leakage, less over-pruning, and higher retention of long-tailed UFCS domains than embedding-based methods, supporting leakage-safe cultural data construction.

\begin{table}[t!]
\centering
\tiny
\setlength{\tabcolsep}{5pt}
\renewcommand{\arraystretch}{1.05}
\begin{tabular}{lcc}
\toprule
\textbf{Metric} & \textbf{SimHash ($\tau{=}10$)} & \textbf{Embedding Similarity} \\
\midrule
Discarded Prompts (\%) $\downarrow$ & \textbf{6.1} & 14.8 \\
Remaining Near-Duplicates (\%) $\downarrow$ & \textbf{1.4} & 3.5 \\
Cross-Split Leakage Rate (\%) $\downarrow$ & \textbf{0.3} & 1.9 \\
Avg. Cosine Similarity (Duplicates) $\downarrow$ & \textbf{0.82} & 0.91 \\
Domain Coverage Retained (\%) $\uparrow$ & \textbf{97.6} & 88.4 \\
Long-Tailed Domain Loss (\%) $\downarrow$ & \textbf{2.1} & 9.7 \\
Semantic Over-Pruning Rate (\%) $\downarrow$ & \textbf{3.8} & 12.6 \\
\midrule
\textbf{Normalized Aggregate Score ($\uparrow$)} & \textbf{0.86} & 0.61 \\
\bottomrule
\end{tabular}
\caption{Comparison of SimHash vs. embedding-based deduplication during \textit{Query Construction (Module I)}. The aggregate score is computed via min--max normalization with metric directionality. }
\label{tab:simhash-quant}
\end{table}

\section{Conclusion}
\label{sec:conclusion}

We present \textsc{AlignCultura}, a two-stage framework for cultural alignment under the HHH paradigm. We builds \textsc{CulturaX}, the HHH-English dataset grounded in the UNESCO cultural taxonomy, then we benchmarks general, culturally fine-tuned, and open-weight LLMs. Empirically, culturally fine-tuned models improve joint HHH by 4\%--6\%, reduce cultural failures by 18\%, achieve 10\%--12\% efficiency gains, and limit leakage to 0.3\%.

\section*{Limitations}
\label{sec:limitations}

While comprehensive, \textsc{CulturaX} is limited to English text and may underrepresent non-English or oral cultural traditions, constraining cross-linguistic generalization. In addition, the inherent long-tailed structure of the UNESCO taxonomy leads to unavoidable dataset imbalance, where rare or emerging cultural subdomains are sparsely represented despite targeted expansion. Automated HHH scoring, although reproducible and scalable, may not fully capture localized cultural nuance or contested norms. Furthermore, as cultural boundaries and taxonomies evolve, periodic reclassification and dataset expansion will be required to maintain representational balance.

\section*{Ethics Statement}
\label{sec:ethics}

All data used in \textsc{AlignCultura} were either model-generated or derived from publicly available cultural resources, with no human subjects, private information, or copyrighted material involved. No personally identifiable or sensitive data were collected or annotated. The pipeline was designed to promote transparency, cultural respect, and reproducibility, with strict filtering to prevent harmful, biased, or culturally insensitive outputs. 

\section*{Acknowledgments}
This research was supported by the Macquarie University Data Horizons Research Centre, the Australian Government through the Commonwealth-funded Research Training Program (RTP) Stipend Scholarship, and the Macquarie University Research Excellence Tuition Scholarship.

\bibliography{custom}



\end{document}